\documentclass[a4paper,fleqn]{cas-sc}
\usepackage[numbers,sort&compress]{natbib}
\usepackage{graphicx}%
\usepackage{multirow}%

\usepackage{xcolor}%
\usepackage{textcomp}%
\usepackage{manyfoot}%
\usepackage{booktabs}%
\usepackage{algorithm}%
\usepackage{algorithmicx}%
\usepackage{algpseudocode}%
\usepackage{listings}%
\usepackage{adjustbox}   % 更安全地控制图像宽度
\usepackage{caption}     % 允许在表格 minipage 中使用 \captionof
\usepackage{booktabs}    % 用于高质量表格线条
\usepackage{subcaption} % 必须加载
\usepackage{array}  % 别忘了加在导言区
\usepackage{capt-of}  % 或者 \usepackage{caption}
\usepackage{etoolbox}
\usepackage{amsmath,amssymb}
\usepackage[utf8]{inputenc}

\usepackage{float}  % 启用[H]参数（强制固定位置）
\usepackage{placeins} % 防止图片跨节乱跑
\usepackage{bm} % 用于更美观的向量表示（\bm{}）
\def\tsc#1{\csdef{#1}{\textsc{\lowercase{#1}}\xspace}}
\tsc{WGM}
\tsc{QE}
\begin{document}
\let\WriteBookmarks\relax

\shorttitle{}    

\shortauthors{}  

\title [mode = title]{SRL: Combining SLIP Model and Reinforcement Learning for Agile Robotic Jumping} 

% ---------------------- 作者信息（CAS模板标准格式） ----------------------
% 第一作者：Xiaowen Hu（上海大学）
\author[1]{Xiaowen Hu}
\fnmark[1]  % 作者脚注标记
\ead{hxw2035143846@shu.edu.cn}  % 邮箱
%\ead[orcid]{0009-0005-3361-5778}
\credit{Conceptualization, Methodology, Software, Formal Analysis, Investigation, Validation, Writing – Original Draft}   % 无贡献标注则留空

% 通讯作者：Linqi Ye（上海大学，加*标记）
\author[1]{Linqi Ye}
\fnmark[2]
\cormark[1]
\ead{yelinqi@shu.edu.cn}
%\ead[url]{https://linqi-ye.github.io/}
\credit{Conceptualization, Supervision, Funding Acquisition, Resources, Project Administration, Writing – Review \& Editing}
 
% 第三作者：Yudi Zhu（上海理工大学）
\author[2]{Yudi Zhu}
\fnmark[3]
\ead{221240082@st.usst.edu.cn}
\credit{Software, Validation, Data Curation}

% 第四作者：Chenyue Shao（上海理工大学）
\author[2]{Chenyue Shao}
\fnmark[4]
\ead{13736413526@139.com}
\credit{Methodology, Software, Formal Analysis}

% 第五作者：Rankun Li（上海大学）
\author[1]{Rankun Li}
\fnmark[5]
\ead{rankunli@shu.edu.cn}
\credit{Investigation, Data Curation}

% 第六作者：Qingdu Li（上海理工大学）
\author[2]{Qingdu Li}
\fnmark[6]
\ead{liqd@usst.edu.cn}
\credit{Supervision, Resources, Validation}

% 第七作者：Yan Peng（上海大学）
\author[1]{Yan Peng}
\fnmark[7]
\ead{pengyan@shu.edu.cn}
\credit{Supervision, Resources, Writing – Review \& Editing}

% ---------------------- 单位信息（CAS模板标准格式） ----------------------
% 单位1：上海大学人工智能研究院
\affiliation[1]{organization={Institute of Artificial Intelligence, Shanghai University},
            addressline={Shangda Road 99}, 
            city={Baoshan District},
            postcode={200444}, 
            state={Shanghai},
            country={China}}

% 单位2：上海理工大学机器智能研究院
\affiliation[2]{organization={Institute of Machine Intelligence, University of Shanghai for Science and Technology},
            addressline={Jungong Road 516}, 
            city={Yangpu District},
            postcode={200093}, 
            state={Shanghai},
            country={China}}

% 通讯作者说明文字
\cortext[1]{Corresponding author}

\begin{abstract}
Robotic jumping is pivotal in applications such as search and rescue and logistics, where crossing obstacles and enhancing mobility efficiency are critical. The Spring-Loaded Inverted Pendulum (SLIP) model leverages simplified spring–mass dynamics that naturally encode biologically plausible hopping motions, yet its performance degrades on irregular terrain due to idealized assumptions regarding contact and joint dynamics. Meanwhile, Reinforcement Learning (RL) can adapt to diverse and complex environments but often requires extensive data from unguided exploration. The complementary strengths of SLIP’s physically grounded baseline and RL’s adaptive capabilities motivate a hybrid framework that overcomes these individual limitations. We therefore propose Spring-loaded Reinforcement Learning (SRL), which integrates SLIP-based feedforward control signals with RL-driven real-time feedback, enabling continuous optimization of robotic jumping. Experimental results demonstrate that SRL can achieve more stable jumps with much less training time than the baseline method, maintaining an average position tracking error below 0.1\,m and velocity tracking errors within ±3\,\% of the target values. Through bipedal and quadrupedal simulations of ground and stair jumping, as well as sim-to-sim and sim-to-real validations, SRL exhibits robust adaptability to various task requirements and environmental complexities, underscoring its potential for real-world deployment.
\end{abstract}


\begin{highlights}
\item A hybrid SRL framework integrating SLIP model and reinforcement learning for legged robot jumping.
\item Six-phase motion modulation for enhanced jumping stability and dynamics.
\item Higher success rate and training efficiency than SLIP-based MPC and RL-only methods.
\item Verified by simulation and real‑world experiments with accurate trajectory tracking.
\end{highlights}

% Keywords

\maketitle

% Main text
\section{Introduction}\label{sec1}
With the growing adoption of robotics in diverse settings—ranging from high-risk rescue missions to warehouse logistics—the ability for robots to perform complex maneuvers, such as jumping\cite{mo2020jumping,koh2015jumping}, has become increasingly critical. Jumping not only enables robots to overcome substantial obstacles, i.e., gullies and elevated platforms, but also significantly enhances mobility and obstacle avoidance in unstructured terrain\cite{yi2024simulating,wang2024bionic,zhao2024design,liu2024single}.

Contemporary robotic jumping control strategies can be broadly categorized into model-driven methods, such as bio-inspired controllers\cite{ribak2020insect,su2017biomimetic,afschrift2023assisting,ezekiel2024bio,elliott2024bio,kabir2024hop,hong2024slip}, the SLIP model\cite{piovan2015reachability,piovan2016approximation,hamzaccebi2024analysis}, and model predictive control (MPC)\cite{park2021jumping,ahn2021online,ji2022concurrent,he2024cdm,xu2025human,fu2024continuous}; and data-driven methods, primarily RL\cite{kober2013reinforcement,tao2024multiobjective}. Model-driven methods, particularly SLIP-based methods, efficiently capture energy storage and release dynamics, enabling physically grounded jumping motions. However, their applicability is often constrained by oversimplified assumptions, such as neglecting ground friction and multi-joint leg dynamics, which limits adaptability to real-world, uneven terrains\cite{hutter2010slip,he2024running,wensing2013control}.

To address these limitations, data-driven methods, such as RL, offer an alternative by autonomously optimizing control policies through interaction with the environment. RL has demonstrated remarkable adaptability across various locomotion tasks. However, its reliance on trial-and-error learning often results in poor sample efficiency and unstable policy exploration due to the lack of physical priors. In robotic jumping, this leads to inefficient motion strategies and increased training costs, hindering real-world deployment.

To bridge this gap, we propose SRL, a hybrid control framework that integrates the SLIP model with RL to achieve both efficiency and adaptability. In the SRL framework, the SLIP model serves as the foundation of feedforward control, leveraging simplified spring–mass dynamics to generate efficient reference trajectories for jumping motions. Additionally, we design an RL-based feedback module, implemented using Proximal Policy Optimization (PPO)\cite{schulman2017proximal,yu2022surprising,zhao2024zsl}, to generate corrective actions that improve stability and robustness in unstructured terrains. The PPO agent continuously processes environmental observations, including body posture, velocity, and foot placement, and outputs feedback corrections to the SLIP-based reference trajectory. The final control signal is obtained by blending the SLIP-generated feedforward commands with RL-derived feedback corrections through a weighted fusion mechanism, balancing predictive efficiency with adaptive flexibility. To ensure precise motion execution, the output control signal is further fine-tuned by a proportional-derivative (PD) controller, which regulates joint torques for smooth and stable jumping. Through this structured integration of model-driven and data-driven control, SRL achieves robust and efficient robotic jumping, effectively mitigating the limitations of both individual methods.

Our contributions are summarized as follows:
\begin{itemize} \vspace{-0.5em}
    \item We propose SRL, a hybrid control framework for robotic jumping that integrates a six-state Finite State Machine (FSM), structured observations, and phase-specific rewards to combine the physical priors of the SLIP model with the adaptability of RL;
    \item We validate SRL through extensive simulations on both biped and quadruped robots, demonstrating superior learning efficiency and stability over various baseline methods;
    \item We further validate SRL through sim-to-sim and sim-to-real experiments, achieving superior trajectory tracking accuracy and demonstrating a real-world biped jumping height of 15\,cm and forward distance of 20\,cm.
\end{itemize}

\section{Related Works}\label{sec2}

The current mainstream methods for achieving robot jumping include several methods. First, bio-inspired control methods mimic animal jumping movements, combined with model predictive techniques to precisely track and optimize trajectories\cite{ahn2021online,su2017biomimetic,ribak2020insect,zhang2017survey,zhang2020biologically,afschrift2023assisting}. Although this approach can realistically replicate natural jumping behaviors, it heavily relies on high-precision control and sensing technology, making it challenging to implement in complex actions\cite{ezekiel2024bio,elliott2024bio,kabir2024hop}. Second, the SLIP model uses a simplified spring and point mass system to effectively capture the energy dynamics of jumping and walking\cite{piovan2015reachability,piovan2016approximation,hamzaccebi2024analysis,garofalo2012walking,shahbazi2016unified,rummel2010robust,xie2021compliant}. Its structure is simple and easy to implement, but due to the model's oversimplification, it struggles to handle precise jump control in various environments\cite{hutter2010slip,he2024running,wensing2013control}. MPC optimizes control strategies by predicting future states, allowing robots to perform jumps in complex environments\cite{park2021jumping,ahn2021online,ji2022concurrent,he2024cdm,xu2025human,fu2024continuous}. While MPC excels in precise control, it has high computational complexity and demands significant real-time performance. Deep Reinforcement Learning (DRL)\cite{kober2013reinforcement,sang2024lunar}, through interaction with the environment, can learn optimal strategies in continuous action spaces, adapting to nonlinear and unknown challenges\cite{bellegarda2024robust,bellegarda2024quadruped,bellegarda2022cpg,peng2018sim,zhou2024stable}. Although DRL has adaptive and self-learning capabilities, the learning process often requires substantial computational resources. 

In the aforementioned methods, the SLIP model has garnered favor among researchers due to its simplicity and effectiveness in capturing the key characteristics of leg dynamics\cite{full1999templates}. Geyer and Saranli\cite{geyer2019gait} strengthened the application of the SLIP model in biped gait generation through their research. Xie et al.\cite{xie2021compliant} proposed Variable Spring-Loaded Inverted Pendulum Model with Finite-sized Foot (VSLIP-FF) with adjustable leg stiffness and ankle joints to enable compliant bipedal walking. He et al.\cite{he2024running} validated the SLIP model's efficacy in quadruped robot running control, achieving stable high-speed locomotion through comprehensive dynamics analysis.

Despite the SLIP model's effectiveness in simulating basic jumping dynamics, its limitations in adapting to complex environments necessitate the integration of other advanced technologies to enhance the practicality and adaptability of robotic systems. In recent years, RL has emerged as a key technology for completing complex tasks in the field of robotics, particularly demonstrating unique advantages in scenarios where specific instructions are not predetermined. Through real-time interaction with the environment and a trial-and-error process, RL enables robots to explore and learn optimal or near-optimal decision-making strategies. This versatility has allowed RL to find broad applications in various fields such as robotic arm control and autonomous navigation\cite{kober2013reinforcement,tao2024multiobjective}. Additionally, through the implementation of the Deep Deterministic Policy Gradient (DDPG) algorithm, RL has demonstrated significant potential in continuous action spaces, effectively facilitating the generation of complex actions such as jumping\cite{bellegarda2024quadruped,bellegarda2024robust}. Moreover, RL has gradually revealed its potential in enabling robotic jumping capabilities\cite{tao2024multiobjective,sang2024lunar}.

To the best of our knowledge, despite the significant progress made in various domains, the integration of the SLIP model with RL to achieve autonomous and efficient jumping in more complex environments remains an underexplored area.

%Table~\ref{Comparison} summarizes the performance comparison between existing methods and our method.

\section{Methods}\label{sec3}

\begin{figure*}
    \centering
    \includegraphics[width=\textwidth]{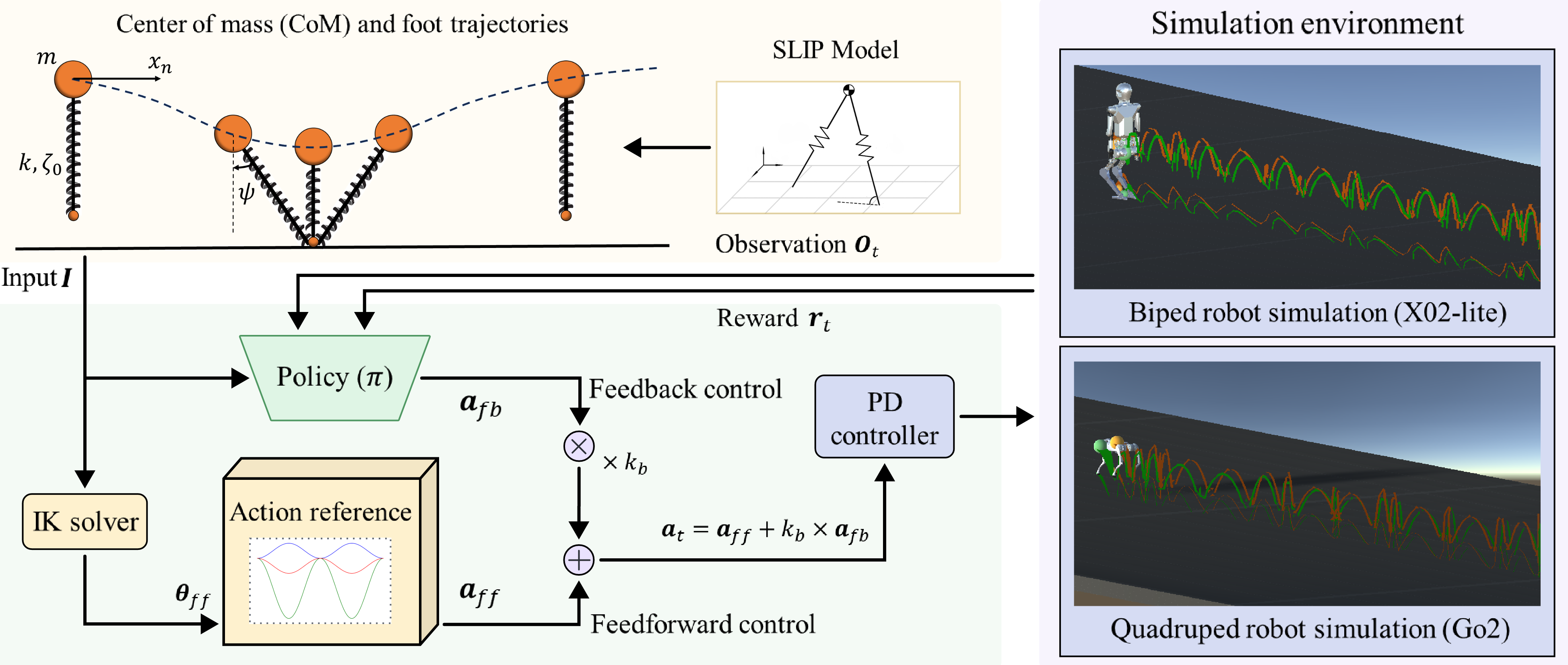}
    \caption{Overview of the SRL control framework for robot jumping tasks, integrating the SLIP model, RL, and simulation environment. SRL combines the physically grounded motion dynamics of the SLIP model with the adaptability of RL to optimize jumping performance in complex environments such as flat ground with varying disturbances, stairs, and boxes.}
    \label{frame}
\end{figure*}

\subsection{SRL Structure}\label{subsec1}
This study proposes a novel robot jumping control framework, SRL, which integrates the SLIP model with RL, as shown in Fig.~\ref{frame}. SRL consists of three core components: 1) the SLIP model, 2) the RL module, and 3) the simulation environment. The SLIP model is responsible for generating feedforward control signals, which are then translated into specific motor commands through an Inverse Kinematics (IK) solver. The RL module continuously refines the control policy by processing environmental observations and reward signals, generating real-time feedback control. The final control signal is a weighted combination of the feedforward and feedback outputs, which is further fine-tuned by a PD controller to achieve precise motion. SRL has been validated in simulations for biped and quadruped robots as well as in selected real-world environments, demonstrating its adaptability and great performance in a variety of jumping tasks.

\subsection{SLIP Model}
\begin{figure*}
    \centering
    \includegraphics[width=0.4\textwidth]{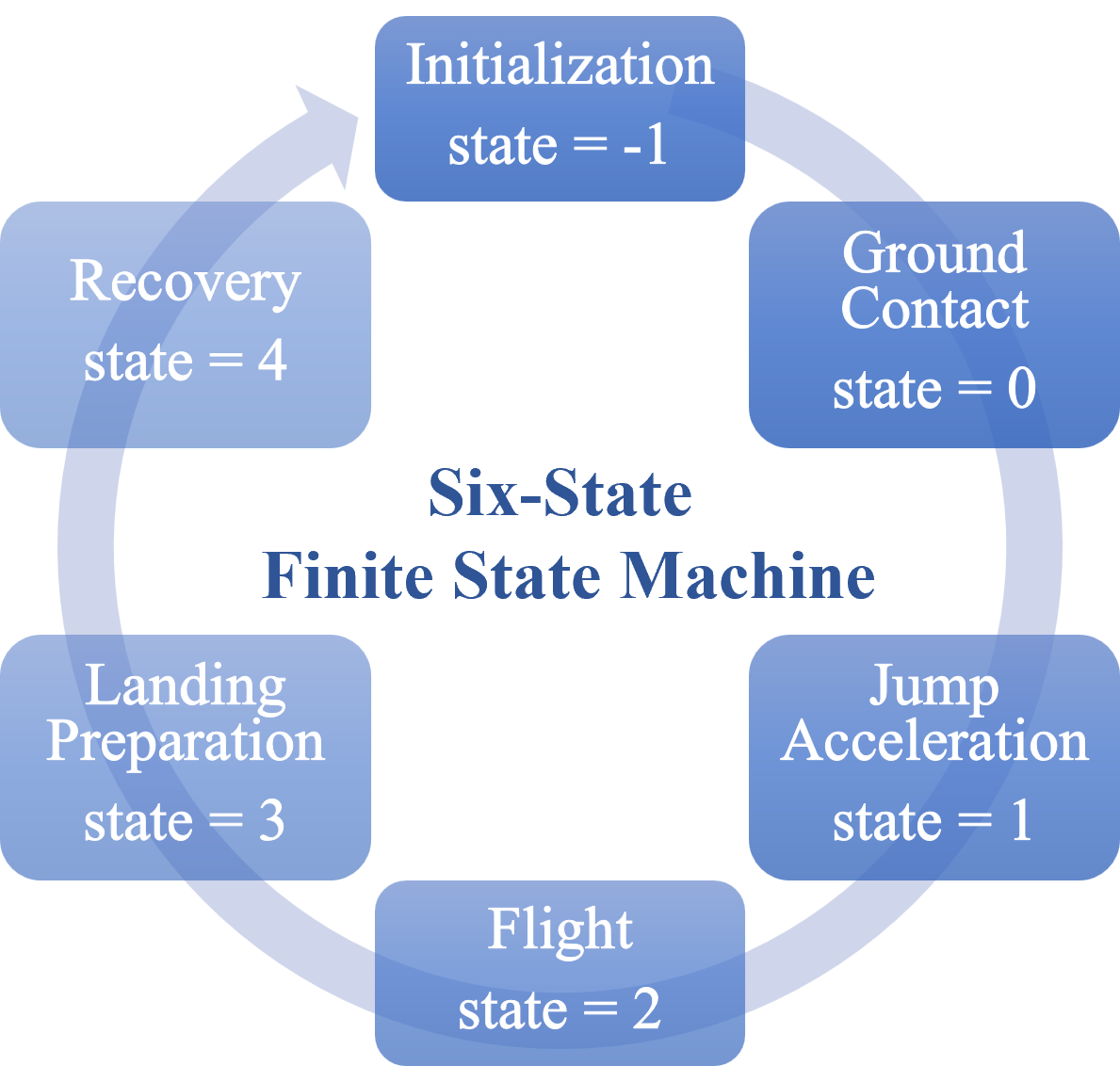}
    \caption{A six-state FSM constructed based on the SLIP model, where each state represents a key stage in the jump cycle.}
    \label{state}
\end{figure*}

The SLIP model, illustrated in the upper left corner of Fig.~\ref{frame}, consists of a point mass connected to a single spring, representing a simplified virtual leg. This model emulates legged locomotion through a dynamic mass-spring system. We developed a six-state FSM (Fig.~\ref{state}) that achieves stable jumping patterns through state-dependent parameter modulation, including dynamic adjustment of spring stiffness, damping coefficients, and rest length across different locomotion phases. The forces on the CoM in each state can be expressed by the equation:
\begin{equation}
F_i = -k_i \Delta L - c_i v_{{m}},
\end{equation}

where $ i \in \{-1,0,1,2,3,4\} $ denotes the state machine mode, 
$ \Delta L = \left( \|\bm{p}_{ {m}} - \bm{p}_{ {f}}\| - \|\bm{p}_{{m}0} - \bm{p}_{ {f}0}\| \right) $ 
represents the change in the distance between the CoM and the foot relative to its initial value, 
$ v_m $ is the velocity of the CoM, 
$ k_i $ and $ c_i $ are the state-dependent stiffness and damping coefficients, respectively.

\subsubsection{Initialization Phase (State -1)}
In the initialization phase, the robot sets the initial positions of the Center of Mass ($ \bm{p}_{ {m}} $) and the foot ($ \bm{p}_{ {f}} $), and calculates the initial force $ F_{\text{-}1} $ based on the displacement between $ \bm{p}_{ {m}} $ and $ \bm{p}_{ {f}} $. Once initialization is complete, the state transitions to the next phase.

\subsubsection{Ground Contact Phase (State 0)}
In the ground contact phase, the robot's foot makes contact with the ground, and the leg compresses like a spring, generating force $ F_0 $ to push the CoM upward. The magnitude of the force is determined by the displacement $ \Delta L $ between the CoM and the foot, and the unit direction vector $ \bm{\hat{n}} $. When the displacement $ \Delta L $ becomes sufficiently small and the vertical component of the velocity of the CoM $(\bm{v}_{ {m}})_y > 0 $, the state transitions to the jump acceleration phase.

\subsubsection{Jump Acceleration Phase (State 1)}
In this phase, the robot further increases the force $ F_1 $ to accelerate the CoM upward, ensuring the robot can leave the ground and enter the flight phase. This phase is primarily responsible for providing the robot with sufficient kinetic energy to complete the jump. When the displacement $ \Delta L $ exceeds a threshold, the state transitions to the flight phase.

\subsubsection{Flight Phase (State 2)}
In the flight phase, the robot leaves the ground and enters free motion. No external forces are applied during this phase. The motion of the CoM ($\bm{p}_{ {m}}$) is dominated by inertia, while the foot position ($\bm{p}_{ {f}}$) is adjusted based on the previous relative displacement to ensure the correct landing position. When $(\bm{v}_{ {m}})_y < 0$ and the foot approaches the ground, the state transitions to the next phase. No force is applied during this phase.

\subsubsection{Landing Preparation Phase (State 3)}
In this phase, the robot begins to gradually contact the ground, applying a cushioning force $ F_3 $ to slow down the descent of the CoM. When the vertical velocity of the CoM $ (\bm{v}_{ {m}})_y $ starts to rise, the robot transitions to the recovery phase. This phase applies a smaller force to cushion the landing.

\subsubsection{Recovery Phase (State 4)}
In the recovery phase, the robot re-establishes contact with the ground, and the force is calculated based on the initial displacement to restore the CoM and foot to their original relative positions. When the positions of the CoM ($\bm{p}_{ {m}}$) and the foot ($\bm{p}_{ {f}}$) approach the initial state, the state transitions back to State -1, initiating a new jumping cycle.

\subsection{Robot IK}
To apply the theoretical SLIP model to an actual robot, we employ an analytical IK method to determine the optimal angles for each joint. Taking a biped robot as an example, the target positions of the robot's CoM ($\bm{p}_{ {m}}$) and the foot ($\bm{p}_{ {f}}$) are obtained according to the SLIP model, from which the corresponding angles of its hip, knee, and ankle joints are calculated.

First, the distance $ d_1 $ between the CoM and the foot is calculated, and the hip joint angle is determined using trigonometric relationships:

\begin{equation}
dc_1 = \arccos \left( \frac{\mathrm{L}_1^2 + d_1^2 - \mathrm{L}_2^2}{2 \mathrm{L}_1 d_1} \right) + \arcsin \left( \frac{(\bm{p}_{ {m}} - \bm{p}_{ {f}})_z}{d_1} \right),
\end{equation}
where  $ \mathrm{L}_1  $ and $ \mathrm{L}_2 $  are the lengths of the thigh and shank, respectively, and $ (\bm{p}_{ {m}} - \bm{p}_{ {f}})_z $ denotes the displacement between the center of mass (CoM) and the foot along the $ z $-axis.

Next, the knee joint angle $ dc_2 $ is computed using the following equation:

\begin{equation}
dc_2 = - \arccos \left( \frac{\mathrm{L}_1^2 + \mathrm{L}_2^2 - d_1^2}{2 \mathrm{L}_1 \mathrm{L}_2} \right) - dc_1.
\end{equation}

Finally, the ankle joint angle $ dc_3 $ is determined by the sum of the hip and knee joint angles:

\begin{equation}
dc_3 = dc_2 + dc_1.
\end{equation}

Through these equations, the IK method enables fast and accurate adjustments to the biped robot's joint angles, ensuring the correct positioning and posture of the foot during the jump.

\subsection{{RL}}
In our previous work\cite{ye2026knowing}, we proposed a learning framework named Instructional Learning, which integrates feedforward and feedback control mechanisms inspired by human learning processes. This method has proven to be efficient, flexible, and generalizable in robot motion learning. In the current research, we build upon this ``Instructional Learning'' method to design an extended control system architecture, integrating it with RL — specifically, PPO~\cite{schulman2017proximal,yu2022surprising,zhao2024zsl} — to optimize the robot's jumping tasks. PPO is a widely-used policy gradient algorithm particularly suited for continuous control tasks such as robotic locomotion. By employing a clipped surrogate objective function, it ensures stable policy updates, preventing overly aggressive changes that could destabilize learning. This enables improved convergence and robust performance in complex environments.

\vspace{0.1cm}
\textbf{Feedforward Signal Design:}
The feedforward signal in our system is derived from the SLIP model, denoted as \( \bm{a}_{ {ff}}(t) = \bm{f}( {SLIP}, t) \), and serves as an initial reference trajectory for the controller. Specifically, the joint angles obtained from IK solver are used as feedforward inputs to the neural network. For the biped robot, this includes the pitch angles of the hip, knee, and ankle joints. In the case of the quadruped robot, only the hip and knee joint angles are used, as the design does not include ankle joints.

\vspace{0.1cm}
\textbf{Action Space:}
The robot's actions are generated by combining the feedforward control signal with the feedback control signal produced by the RL algorithm. At each time step, the joint control signal is calculated. The weighted sum of the feedback signal \( \bm{a}_{ {fb}} \) and the feedforward signal \( \bm{a}_{ {ff}} \) forms the final target joint angle configuration: \( \bm{a}_t = \bm{a}_{ {ff}} + k_b \times \bm{a}_{ {fb}}
 \). Finally, a PD controller is used to convert this target angle into joint torques, ensuring the precise execution of the jumping motion.

\vspace{0.1cm}
\textbf{State Space:}
In our state space \( \bm{O}_t \), the agent observes key dynamic parameters of the robot during the jumping process, enabling precise adjustment of its movements. The state space includes the following observations: the robot’s body linear velocity \( \bm{v}_{ {base}} \) and angular velocity \( \bm{\omega}_{ {base}} \) in the local coordinate frame, as well as the pitch angle \( \theta_{ {pitch}} \) and roll angle \( \theta_{ {roll}} \). Additionally, the joint angle deviations \( \Delta a_1, \Delta a_2, \dots, \Delta a_n \), which represent the difference between the feedforward signal (the target angles from the SLIP model) and the actual joint angles, are also included. Joint velocities \( v_1, v_2, \dots, v_n \) are further observed to monitor the dynamic state of the joints. These observations provide the agent with a comprehensive understanding of the robot’s posture and movement, allowing it to adjust actions dynamically to maintain stability and precision during complex jumping tasks.

\vspace{0.1cm}
\textbf{Reward:} In our robot jumping task, the reward function is crucial to ensure precise and stable movement control. The specific implementation of the reward function is described through the following components:

1) \textbf{Live Reward:}  
This reward ensures that the robot remains active during each time step, which is essential for completing long-term jumping tasks.  
\begin{equation}
r_{ {live}} = 1.
\end{equation}

2) \textbf{Orientation Reward:}  
This reward penalizes deviations in pitch, yaw, and roll angles to ensure the robot's stability in mid-air and upon landing.
\begin{equation}
r_{ {ori}} = -k_p \cdot |\theta_{ {pitch}}| - k_y \cdot |\theta_{ {yaw}}| - k_r \cdot |\theta_{ {roll}}|,
\end{equation}
where  
\( \theta_{ {pitch}} \) is the pitch angle deviation,  
\( \theta_{ {yaw}} \) is the yaw angle deviation,  
\( \theta_{ {roll}} \) is the roll angle deviation,  
\( k_p \), \( k_y \), \( k_r \) are coefficients for the pitch, yaw, and roll angles, respectively.

3) \textbf{Velocity Reward:}  
This reward ensures the robot maintains smooth motion during jumps by penalizing deviations between its actual and expected speed.
\begin{equation}
r_{ {vel}} = -\| \bm{v}_{ {actual}} - \bm{v}_{ {expected}} \|,
\end{equation}
where \( \bm{v}_{ {actual}} \) is the robot's current velocity and \( \bm{v}_{ {expected}} \) is the expected velocity for the task.

4) \textbf{Synchronization Reward:}  
This reward penalizes differences in joint angles between the legs, ensuring synchronized movement.  

For biped robots:  
\begin{equation}
r_{ {syn}} = - \sum_{i=1}^{N} \lvert a_{i}^{ {left}} - a_{i}^{ {right}} \rvert.
\end{equation}

For quadruped robots:  
\begin{equation}
r_{ {syn}} = - \sum_{i=1}^{N} \left( \lvert a_{i}^{ {FL}} - a_{i}^{ {FR}} \rvert + \lvert a_{i}^{ {RL}} - a_{i}^{ {RR}} \rvert \right).
\end{equation}

5) \textbf{Foot Position Tracking Reward:}  
This reward minimizes the deviation of the robot's foot position from the target position, ensuring that each step follows the predetermined trajectory.
\begin{equation}
r_{ {pos}} = - \sum_{j=1}^{M} \| \bm{p}^{ {actual}}_{j} - \bm{p}^{ {target}}_{j} \|,
\end{equation}
where \( \bm{p}^{ {actual}}_{j} \) is the actual foot position of leg \( j \), and \( \bm{p}^{ {target}}_{j} \) is the target foot position.

6) \textbf{Total Reward:}  
The total reward is a weighted combination of all the individual rewards.
\begin{equation}
r_{ {total}} = r_{ {live}} + w_1 r_{ {ori}} + w_2 r_{ {vel}} + w_3 r_{ {syn}} + w_4 r_{ {pos}},
\end{equation}
where \( w_1 \), \( w_2 \), \( w_3 \), and \( w_4 \) are the weight coefficients for the orientation, velocity, synchronization, and foot position rewards, respectively. The control parameters and reward weights are summarized in Table~\ref{key_parameters}.

\begin{table}[htbp]
\centering
\caption{Key Control and Reward Parameters of the Proposed SRL Framework}
\label{key_parameters}
\begin{tabular}{ccc|ccc}
\toprule
\textbf{Parameter} & \textbf{Biped} & \textbf{Quadruped}
& \textbf{Reward Weight} & \textbf{Early Stage} & \textbf{Late Stage} \\
\midrule
$k_b$ & 30 & [10, 30, 60]$^\dagger$
& $w_1$ & 2 & 1 \\

$k_p$ & 0.1 & 0.02
& $w_2$ & 0.5 & 1 \\

$k_y$ & 0.5 & 2
& $w_3$ & 0 & 1 \\

$k_r$ & 0.1 & 0.02
& $w_4$ & 1 & 1 \\
\bottomrule
\end{tabular}

\vspace{1mm}
\footnotesize{$^\dagger$ For the quadruped robot, per-leg feedback gains are 10, 30, and 60 for the hip, thigh, and knee joints, respectively.}
\end{table}

Termination Condition: We terminate the episode if the robot ``falls'', defined as either the pitch angle \( \theta_{ {pitch}} \) or the roll angle \( \theta_{ {roll}} \) exceeding 30 degrees. The episode also ends if the time step reaches 1000.

\textbf{Network and Hyperparameters:}
The network architecture and training hyperparameters used by the PPO algorithm are shown in Table~\ref{network_architecture} and Table~\ref{ppo_hyperparams}, respectively.

\begin{table}[htbp]
\centering
\begin{minipage}{0.48\textwidth}
\centering
\caption{Network Architecture for Biped and Quadruped Robots}
\begin{tabular}{c c}
\toprule
\textbf{Hyperparameter}       & \textbf{Value}   \\
\midrule
Actor Hidden Layer            & [512, 512, 512]         \\
Critic Hidden Layer           & [512, 512, 512]         \\
Input Observation Dimension   & 28 / 32 \\
Output Action Dimension       & 10 / 12 \\
Activation Function           & ReLU \\
\bottomrule
\end{tabular}
\label{network_architecture}
\end{minipage}
\hfill
\begin{minipage}{0.48\textwidth}
\centering
\caption{PPO Hyperparameters for Biped and Quadruped Robots}
\begin{tabular}{c c}
\toprule
\textbf{Hyperparameter}       & \textbf{Value}   \\
\midrule
Discount Factor               & 0.995                       \\
GAE Discount Factor           & 0.95    \\
Minibatch Size                & 2024 / 2048                 \\
Timesteps per Rollout         & 20240 / 20480               \\
Epochs per Rollout            & 3                           \\
Entropy Bonus Coefficient     & 0.005                       \\
Value Loss Coefficient        & 1.0             \\
Clip Range                    & 0.2                         \\
Learning Rate                 & 0.0003                      \\
Environments                  & 32                          \\
\bottomrule
\end{tabular}
\label{ppo_hyperparams}
\end{minipage}
\end{table}

\section{Experiments}
\subsection{Simulation Setup}
In this study, we use the Unity engine and ML-Agents toolkit for simulation and training experiments. The simulation runs with a time step of 0.01 seconds, corresponding to a control frequency of 100Hz. To accelerate the training process, we deploy 32 agent instances in parallel, ensuring efficient policy learning.

We selected two types of robots to validate our approach: the X02-lite biped robot, developed by Shanghai Droid Robotics, and the Unitree Go2 quadruped robot, developed by Hangzhou Unitree Technology. These robots represent typical biped and quadruped designs, making them well-suited to evaluate the generalization of SRL. 

\subsection{Training Process}
\subsubsection{Phase-Based Training Strategy}
\begin{figure*}
    \centering
    \includegraphics[width=\textwidth]{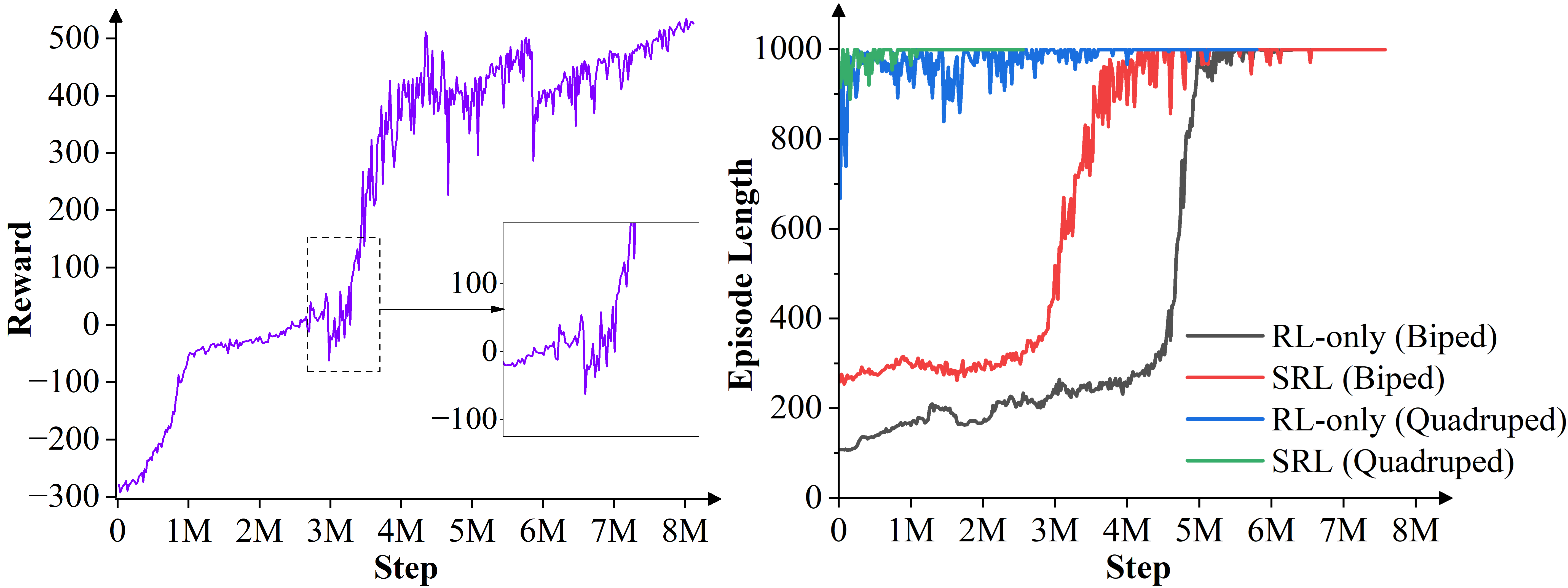}
    \caption{Left: Reward evolution for the biped robot's random-distance jump; Right: Learning efficiency comparison of RL-only and SRL in fixed-distance jumping.}
    \label{reward}
\end{figure*}

The training process is structured in phases to gradually shift the focus from stability to speed and efficiency. For instance, during the training process for the biped robot's random-distance jumping task, the reward function's weight parameters \( w_1 \), \( w_2 \), \( w_3 \), and \( w_4 \) are adjusted across different stages, as shown in the left part of Fig.\ref{reward} and summarized in Table~\ref{key_parameters}. Initially, the emphasis is placed on stability by assigning a higher weight to the orientation reward. As training progresses, the reward weights are gradually adjusted to prioritize velocity over orientation, aiming to improve the robot's speed and jumping efficiency while maintaining basic stability. This staged reward shaping strategy enables the policy to transition from stable locomotion to speed-optimized performance across diverse jumping distances. 

\subsubsection{Jumping Tasks Design}
We conducted fixed-distance and random-distance jumping tasks with biped and quadruped robots. Fixed-distance tasks (e.g., 0.1\,m, 0.2\,m) steadily improved strategy performance with continuous goal training, while random-distance tasks (0-0.3\,m for biped robots) may require sudden adaptation to long-distance jumps (e.g., 0.3\,m) after consecutive short-distance jumps, thus rigorously testing the dynamic adaptability of the strategies. Biped training prioritises balance control during jumps because of their high CoM and the need to simultaneously maintain stability and adjust motion during distance changes, in contrast to quadruped robots that exhibit higher stability, which motivated us to perform more challenging tasks for quadruped robots. Therefore we conducted the task of jumping onto a box, which introduces vertical height variations alongside horizontal jumps, comprehensively validating the robustness of the SRL framework in multidimensional motion scenarios.

\subsection{Performance Evaluation}
\subsubsection{Baseline Comparison}
Table~\ref{performance} presents a comprehensive comparison of the learning efficiency across three different control methods: SLIP-based MPC, RL-only(PPO), and Our SRL. Due to fundamental differences in reward structures, a direct comparison of reward values between SRL and RL-only methods is not meaningful. Therefore, we focus on higher-level performance metrics — namely, training steps and success rate — to enable a more valid and insightful evaluation of learning efficiency. Training steps refer to the total number of learning updates necessary to reach a converged policy, defined as consistently achieving episode lengths exceeding 950 and stable reward levels, under a maximum episode length of 1000. Success rate is calculated as the ratio of successful jumps without falling over 1000 test trials (e.g., a success rate of 0.998 corresponds to 998 successful trials).

The SLIP-based MPC method achieves moderate performance with success rates of 68.0\% for the biped and 75.2\% for the quadruped. While this demonstrates the effectiveness of incorporating biomechanical priors into control, the lack of online learning limits its robustness and adaptability. In contrast, the RL-only method demonstrates significantly improved performance, achieving high success rates of 91.3\% for biped and 95.7\% for quadruped, albeit at the cost of substantial training time (7.2M and 3.8M steps respectively). Our proposed SRL method further improves upon these results, attaining the highest success rates — 98.5\% for the biped and 99.8\% for the quadruped — while requiring notably fewer training steps (5.7M and 1.1M steps respectively), as shown in the right part of Fig.~\ref{reward}. This reduction in training complexity suggests that the integration of analytical priors with RL not only enhances performance but also accelerates learning convergence significantly.

\begin{table}[htbp]
\caption{Baseline Comparison and Ablation Study}
\centering
\setlength{\tabcolsep}{2.0mm} % 列间距适配，不拥挤
\begin{tabular}{l c c c c}
\toprule
\textbf{Method} & \multicolumn{2}{c}{\textbf{Training Steps ($10^6$)}} & \multicolumn{2}{c}{\textbf{Success Rate}} \\
\cmidrule(lr){2-3} \cmidrule(lr){4-5}
& Biped & Quadruped & Biped & Quadruped \\
\midrule
\multicolumn{5}{l}{\textbf{(a) Baseline}} \\
SLIP-based MPC   & N/A   & N/A   & 0.680 & 0.752 \\
RL-only (PPO)     & 7.2   & 3.8   & 0.913 & 0.957 \\
Ours (SRL)        & \textbf{5.7} & \textbf{1.1} & \textbf{0.985} & \textbf{0.998} \\
\midrule
\multicolumn{5}{l}{\textbf{(b) Ablations}} \\
Ours   & 5.7   & 1.1   & 0.985 & 0.998 \\
w/o FSM Structure & 7.5   & 3.8   & 0.803 & 0.886 \\
w/o Velocity Reward & 6.2 & 1.6  & 0.895 & 0.912 \\
\bottomrule
\end{tabular}
\label{performance}
\end{table}

\subsubsection{Ablation Study}\label{subsubsec:ablation}

To verify the contribution of each core component in the SRL architecture, we conduct ablation studies by removing the six-state FSM structure and the velocity reward. All variants are trained and evaluated under identical experimental settings. As shown in Table~\ref{performance}, removing either component leads to increased training steps and degraded success rates, highlighting the importance of both phase-aware control and velocity tracking for efficient and stable jumping.

In the “w/o FSM” setting, the discrete phase-transition mechanism is disabled. The jumping cycle is instead generated by a predefined cosine-based oscillator, producing a smooth periodic signal with fixed period and continuous feedforward joint references for bounding locomotion. This time-driven formulation replaces state-dependent phase switching with a fixed-frequency rhythmic pattern. Without adaptive phase modulation, the controller relies on a fixed rhythmic signal, which degrades phase coordination and locomotion stability, resulting in inferior jumping performance.

Removing the velocity reward slows convergence and reduces stability, indicating that velocity tracking provides essential guidance for robust control.

Quantitatively, removing the FSM structure increases training steps from 5.7M to 7.5M and reduces success rate from 98.5\% to 80.3\%, while removing the velocity reward leads to 6.2M training steps and 89.5\% success rate. These results demonstrate that both phase-aware control and velocity tracking significantly improve sample efficiency and final performance, fully validating the proposed SRL framework.

\begin{figure}
    \centering

    \begin{subfigure}{0.9\textwidth}
        \centering
        \includegraphics[width=\linewidth]{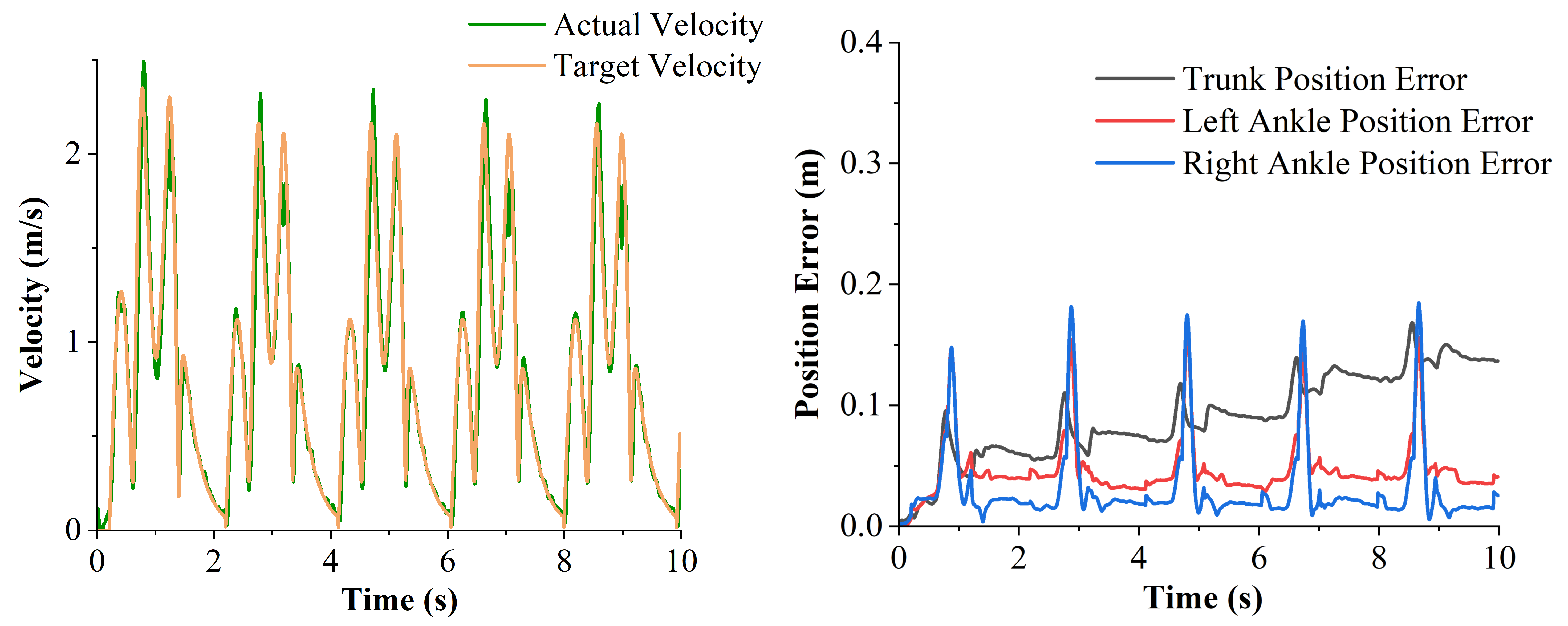}
        \caption{Fixed-distance jumping.}
        \label{fig:bipedvea}
    \end{subfigure}

    \vspace{0.5em}

    \begin{subfigure}{0.9\textwidth}
        \centering
        \includegraphics[width=\linewidth]{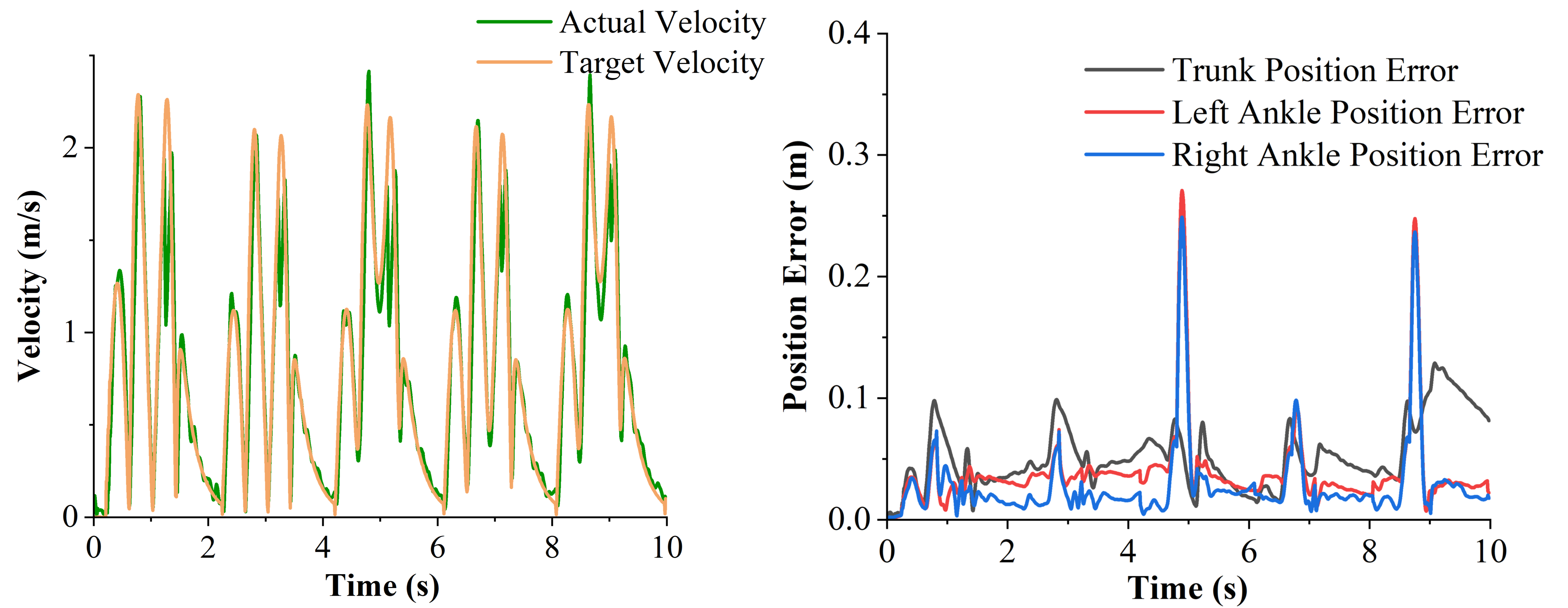}
        \caption{Random-distance jumping.}
        \label{fig:bipedveb}
    \end{subfigure}

    \caption{Performance of the biped robot during fixed-distance jumping, showing trunk velocity, absolute tracking errors of the trunk position, and relative tracking errors of the ankle.}
    \label{fig:bipedve}
\end{figure}

\begin{figure}
    \centering

    \begin{subfigure}{0.9\textwidth}
        \centering
        \includegraphics[width=\linewidth]{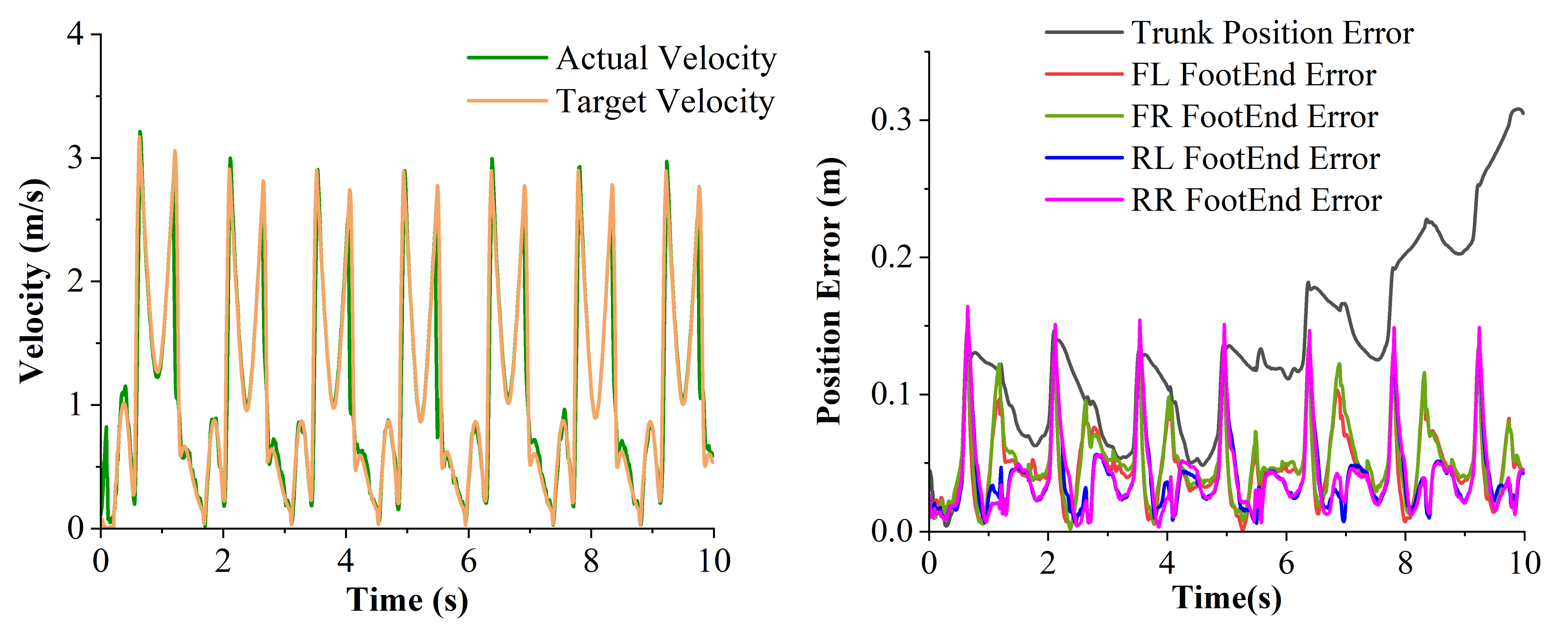}
        \caption{Fixed-distance jumping.}
        \label{fig:quadvea}
    \end{subfigure}

    \vspace{0.5em}

    \begin{subfigure}{0.9\textwidth}
        \centering
        \includegraphics[width=\linewidth]{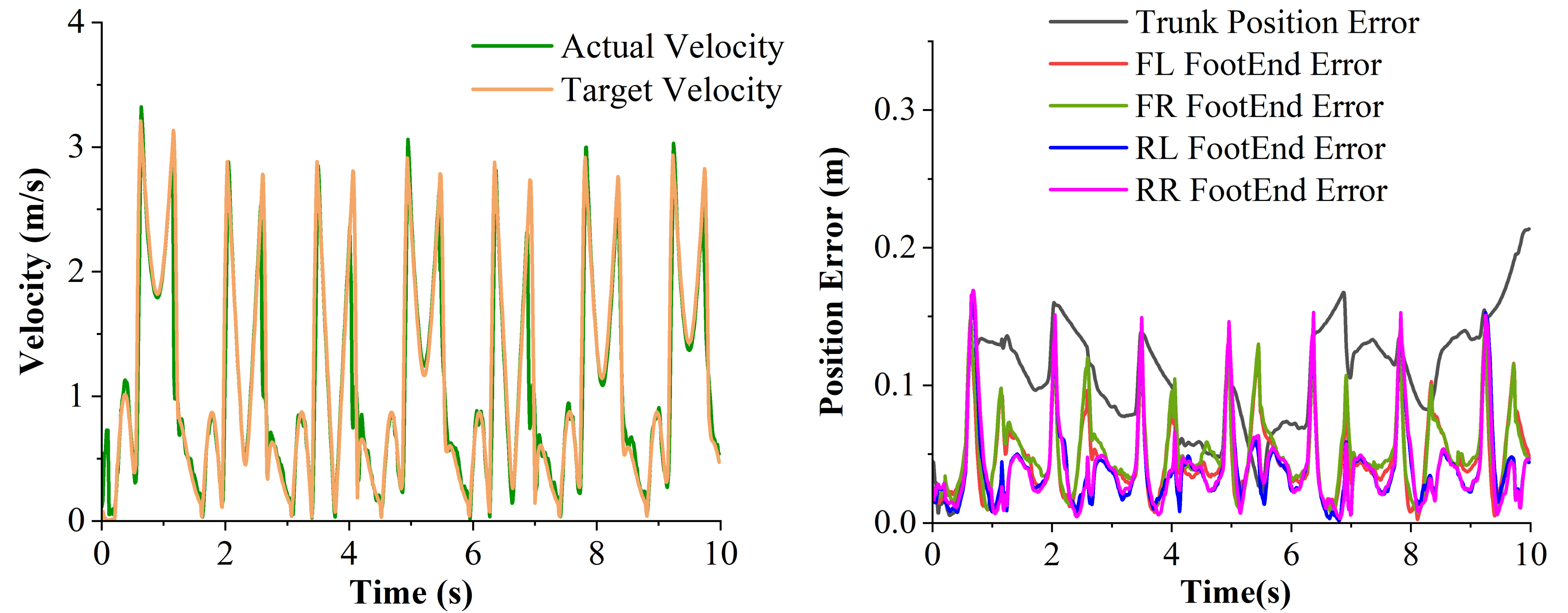}
        \caption{Random-distance jumping.}
        \label{fig:quadveb}
    \end{subfigure}

    \caption{Performance of the quadruped robot during fixed-distance and random-distance jumping tasks, showing trunk velocity, absolute tracking errors of the trunk position, and relative tracking errors of the foot.}
    \label{fig:quadve}
\end{figure}

\begin{figure}
    \centering
    \includegraphics[width=0.9\textwidth]{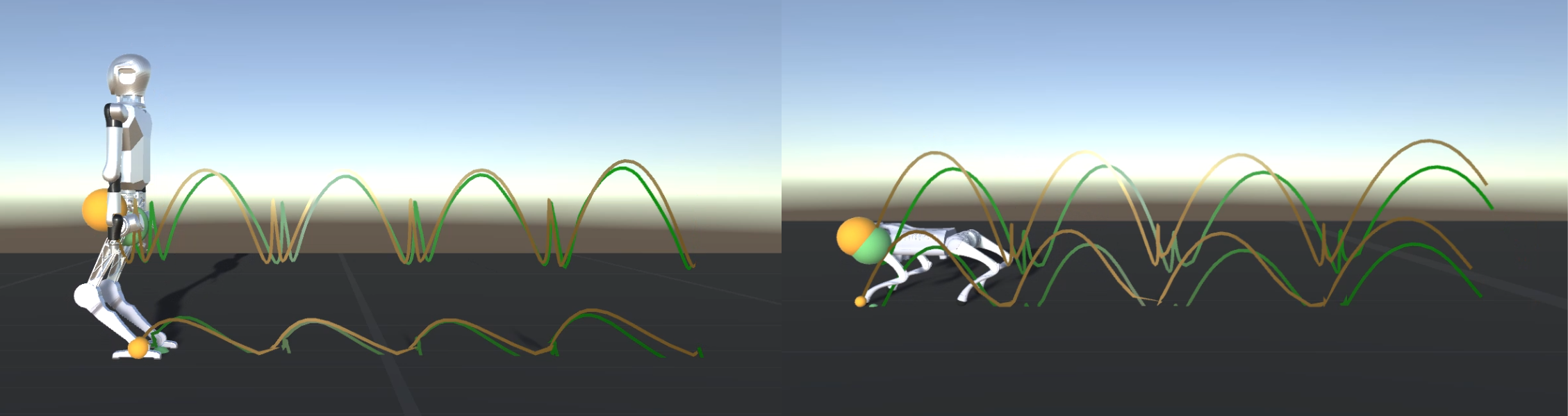}
    \caption{Fixed-distance jumping simulations of the biped and quadruped robots. Orange and green denote the target and actual states, respectively.}
    \label{bipedquad0}
\end{figure}

\subsection{Simulation Results}
\textbf{Biped Robot :} As illustrated in Fig. \ref{fig:bipedve}, the performance of the biped robot during both fixed-distance and random-distance jumping tasks is presented. The figure shows the comparison between the robot's body velocity and the target velocity, as well as the tracking errors of the body and foot-effectors. In the fixed-distance jumping task, the robot effectively tracked the target velocity, reaching a peak speed of 2\,m/s. However, over time, the position tracking errors began to accumulate. Conversely, during the random-distance jumping task, the variation in jump distances allowed the robot to correct its position-tracking errors more easily, though this resulted in a slight reduction in position-tracking accuracy. Overall, across both tasks, the biped robot demonstrated strong capability in tracking the target velocity, keeping body tracking errors within 0.2\,m over a 10-second period. 

Although the foot-effectors occasionally exhibited periodic error spikes during jumping, analysis showed that these peaks primarily occurred at the moment of takeoff and landing, especially immediately after takeoff, when the robot’s leg joints experienced maximum explosive force and its body velocity reached its peak. Due to the system's increased dynamic sensitivity, even small position errors were amplified. Despite the relatively large error spikes shown in the graph, these errors were transient in nature (lasting less than 100 milliseconds), occurring during the flight phase without ground contact and not leading to long-term instability. Rapid adjustments during other phases successfully compensated on average for these errors, maintaining errors within the range of 0 to 0.1\,m. Similar phenomena were observed with quadruped robots performing jumping tasks (Fig.~\ref{fig:quadve}).

\begin{figure*}
    \centering
    \includegraphics[width=0.7\textwidth]{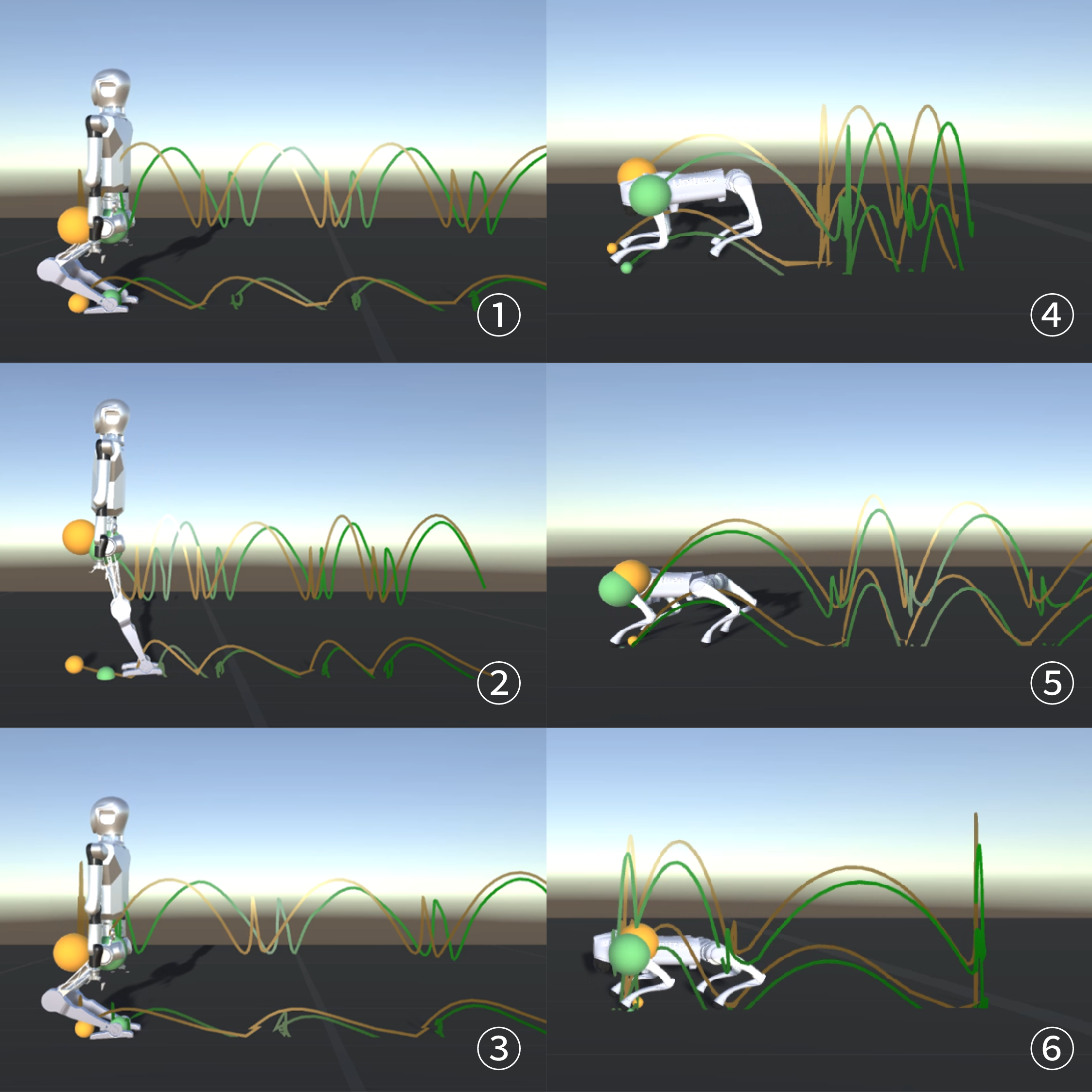}
    \caption{Simulation of the biped (left) and quadruped (right) robot in random-distance jumping tasks, illustrating the motion trajectories of the body and ankle/foot.}
    \label{bipedquad1}
\end{figure*}

Fig.~\ref{bipedquad0} illustrates the body and ankle trajectories of the biped robot during the fixed-distance jumping task, clearly presenting the robot's movement patterns throughout the task. Fig.~\ref{bipedquad1} demonstrates the performance of the biped robot in random-distance jumping tasks. The figure provides a clear visualization of the robot's ability to flexibly adapt to varying jump distances, showcasing its remarkable agility and jumping capabilities. This dynamic adaptability to varied commands further demonstrates the effectiveness and precision of SRL in handling complex and variable conditions.

\textbf{Quadruped Robot :} Compared to the biped robot, the quadruped robot was able to complete a greater number of jumps within the same time frame and achieved higher body velocities, with a maximum speed of 3\,m/s. Similar to the biped case, the quadruped robot demonstrated better velocity-tracking ability during the fixed-distance jumping task but also exhibited an accumulation of position-tracking errors over time. Detailed results can be seen in Fig. \ref{fig:quadve}.

\begin{figure*}
    \centering
    \includegraphics[width=0.9\textwidth]{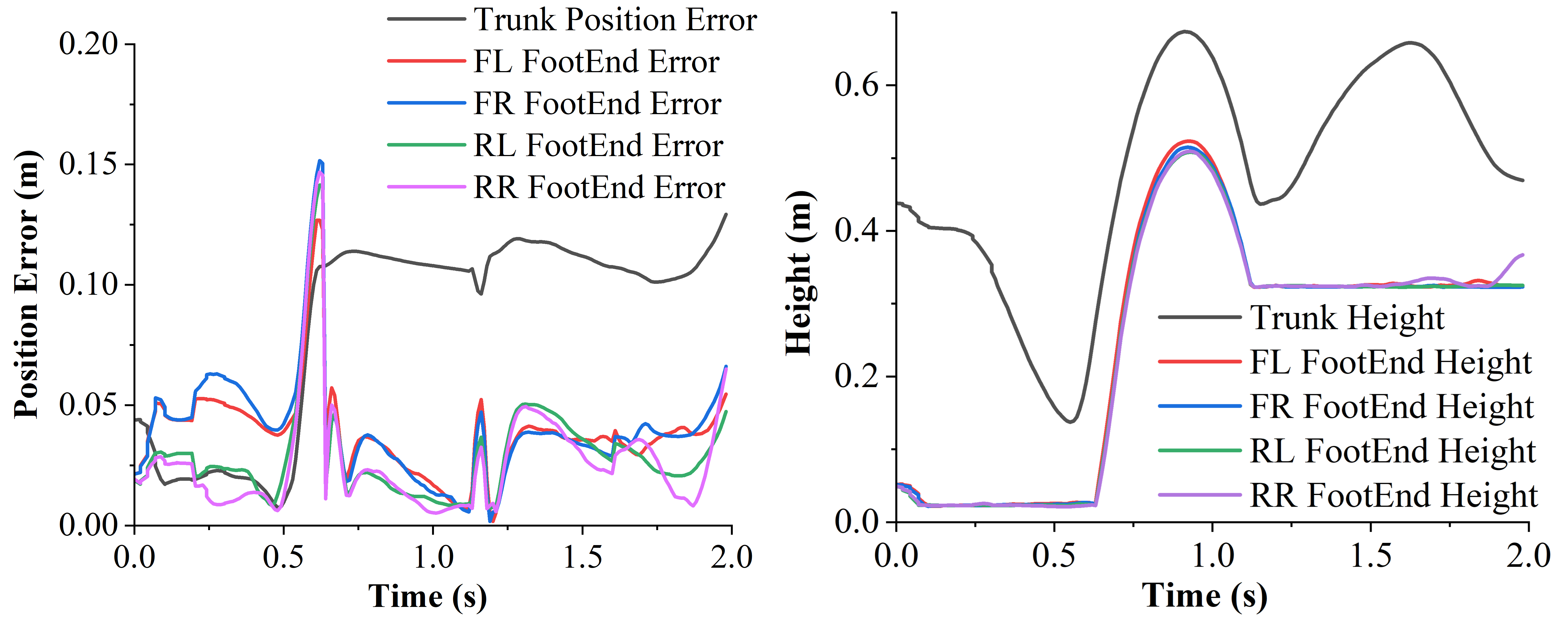}
    \caption{Performance of the quadruped robot during the box-jumping experiment. The left shows absolute body position errors and relative foot errors, while the right shows the height of the body and feet.}
    \label{squadve}
\end{figure*}

The simulation results of the quadruped robot in fixed-distance jumping tasks are shown in Fig.\ref{bipedquad0}. In the random-distance jumping tasks (as shown in Fig. \ref{bipedquad1}), the robot similarly demonstrates the ability to adapt flexibly to varying jump distances, exhibiting excellent stability and jumping efficiency.

\begin{figure*}
    \centering
    \includegraphics[width=0.9\textwidth]{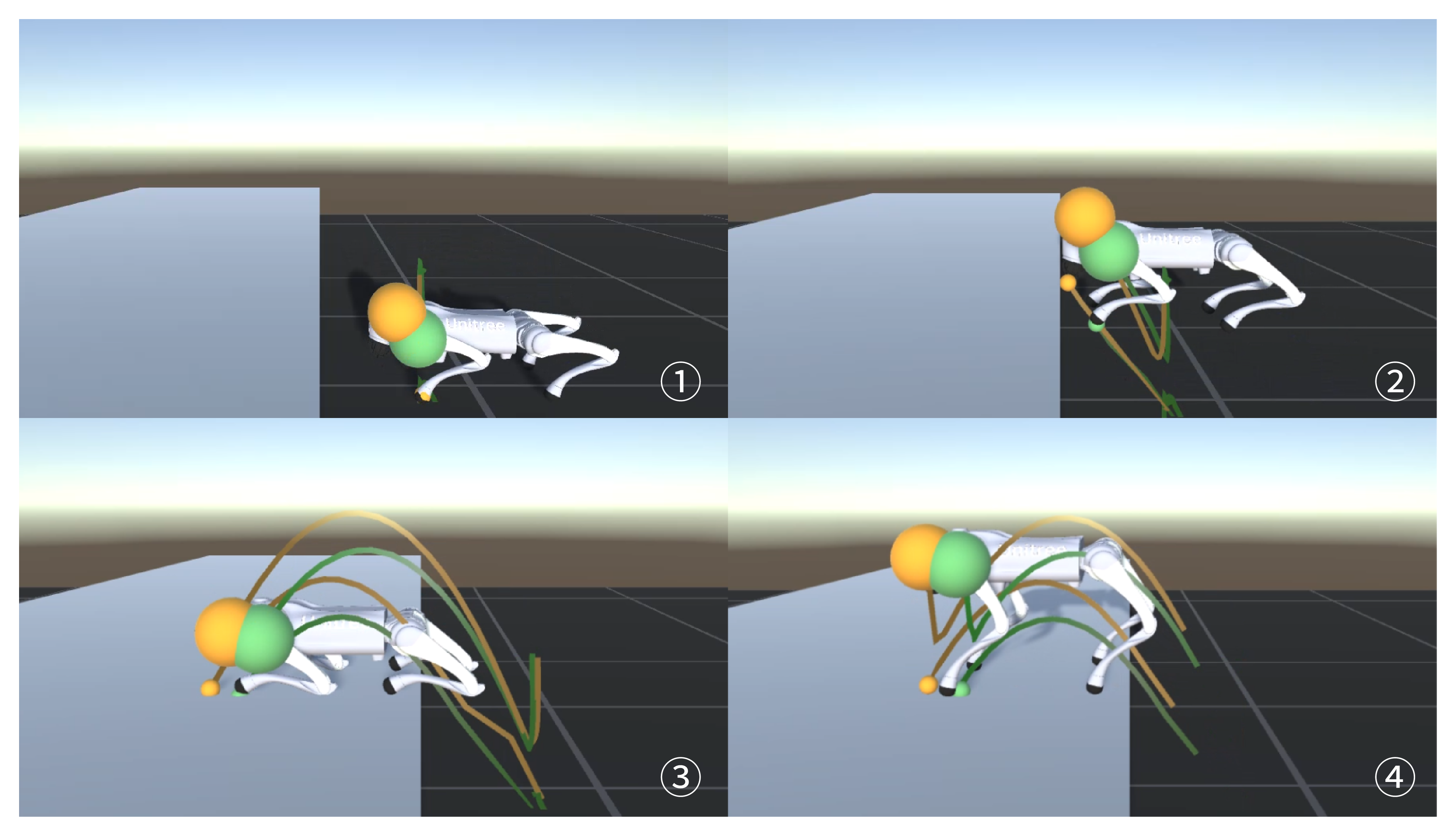}
    \caption{Simulation of the quadruped robot during the box-jumping experiment.}
    \label{squad}
\end{figure*}

%\begin{figure}[htbp]
%    \centering
%    \includegraphics[width=\linewidth]{quadsim.png}
%    \caption{Simulation of the quadruped robot in Isaac Gym.}
%   \label{quadsim}
%\end{figure}

In the box-jumping experiment, the quadruped robot performed great (as shown in Fig. \ref{squadve}). Although the absolute position errors of the body increased over time due to the complexity of synchronizing the four legs to adapt to uneven terrain, the robot was still able to maintain the relative foot errors mostly within 0.05\,m. Moreover, the height (y-coordinate) of the body and feet during the jumping process indicates that the robot successfully adjusted its posture and extended its legs to manage the vertical displacement when stepping up. The trajectories of the four legs were largely synchronized, and the figure clearly shows that the robot was able to successfully jump onto a box with a height of 0.35\,m. Fig.~\ref{squad} shows the simulation with the body and foot trajectories during the box-jumping process, providing a clear visualization of the robot’s coordination and movement patterns throughout the task. 
%In addition, we conducted the same experiment on the quadruped robot on the Issac Gym platform, and the experimental results (Fig.~\ref{quadsim}) also validated the feasibility and adaptability of our method.

The above results confirm the effectiveness of our SRL framework across various locomotion tasks and platforms. It is worth noting that the current policies are trained separately for each specific task, given the distinct control demands across different jumping scenarios. Exploring a unified policy that generalizes across diverse tasks through high-level commands would be an important direction for future work.

\subsection{Real-World Experiments}
\begin{figure*}
    \centering
    \includegraphics[width=0.8\textwidth]{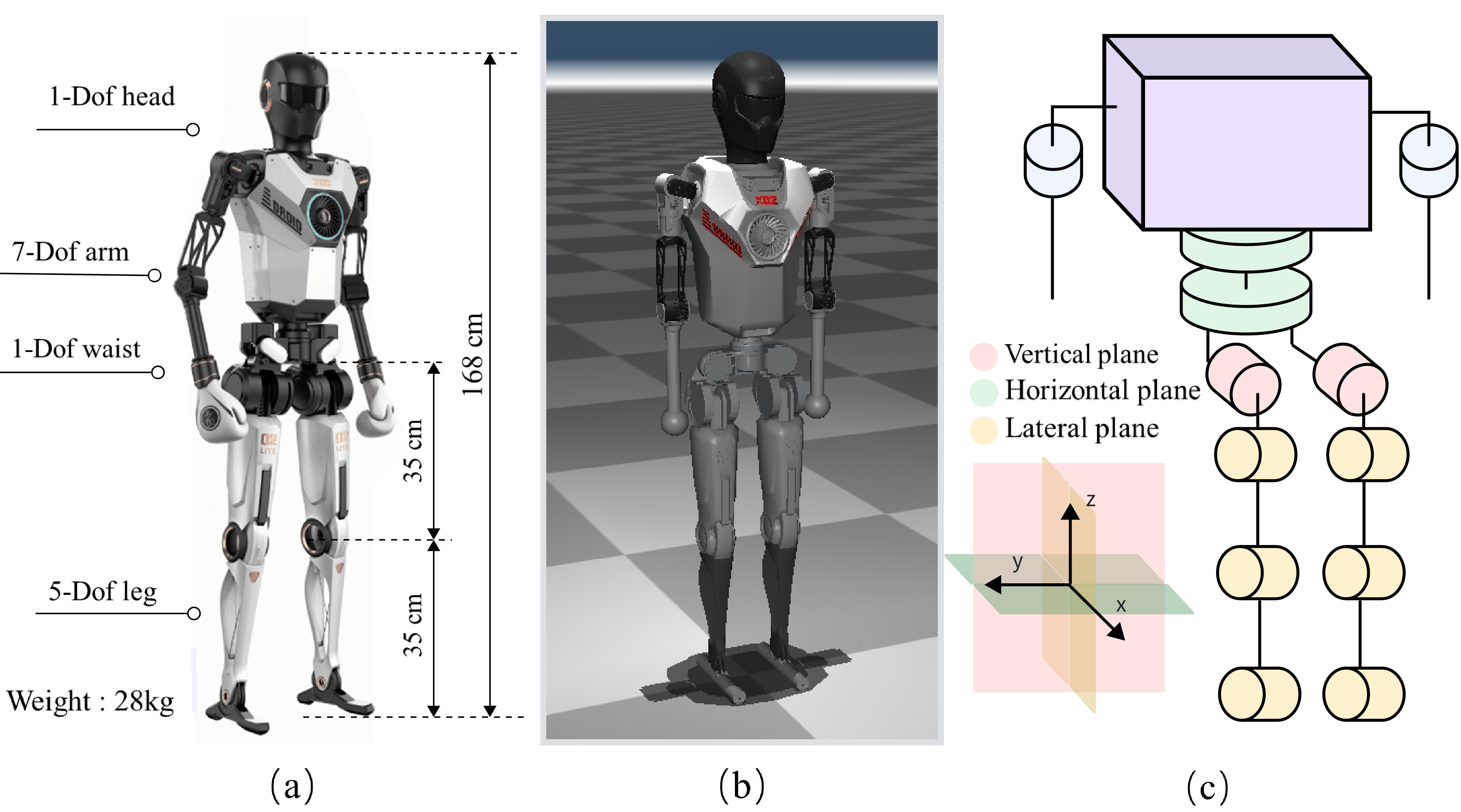}
    \caption{The X02-lite humanoid robot (Shanghai Droid Robotics) used in the experiment, featuring a height of 168\,cm and a total mass of 28\,kg. (a) Real robot. (b) Mujoco model. (c) Simplified link model (focusing on the 10 DOF legs).}
    \label{x02}
\end{figure*}

\begin{table}
\caption{Domain Randomization Parameters}
\centering
\begin{tabular}{c c}
\toprule
\textbf{Parameter}                 & \textbf{Range[Min,Max]}       \\
\midrule
Base Mass        &  [-3, 3] kg  \\
CoM           &  [-0.15, 0.15] m \\
Ground Friction          & [0.5, 1.5]  \\
Restitution   & [0.2, 1.0]     \\
Push Interval              & 4 s              \\
Push Velocity (XY)    & [0, 0.5] m/s           \\ 
Push Angular Velocity & [0, 0.4] rad/s      \\
Motor Strength      & [0.9, 1.2] $\times$ default Nm      \\ 
Motor Offset       & [-0.05, 0.05] rad $\times$ default rad \\ 
Joint Kp           & [0.8, 1.3]$\times$ default   \\ 
Joint Kd         & [0.5, 1.5]$\times$ default      \\ 
Gravity Interval & 7 s                              \\ 
Gravity Impulse Duration  & 1.0 s                                \\ 
Gravity Range     & [-0.5, 0.5] $\times$ default m/s²                     \\ 
\bottomrule
\end{tabular}
\label{domain_rand_params}
\end{table}

To further validate the proposed SRL framework, we conducted real-world experiments using a biped robot platform and compared the results with our simulation outcomes. The physical robot used in our experiments is the X02-lite biped robot (Fig. \ref{x02}).

X02-lite has a total height of 168\,cm and an arm span of 167\,cm, closely resembling average human proportions. It weighs 28\,kg and features a lower body with two 35\,cm leg segments — thigh and shank. The system includes 26 Degrees of Freedom (DOF), with each leg having 5 independently actuated joints, including a knee joint capable of delivering up to 180\,Nm of peak torque. Joint positions are measured using a dual-encoder configuration, providing high-precision feedback for motion control. Low-level control runs at 1\,kHz for fast and responsive joint dynamics. Environmental perception is achieved through a stereo vision system and a YIS320 IMU, which combines triaxial MEMS accelerometers and gyroscopes for real-time 3D orientation estimation. To better utilize these sensor measurements and estimate linear velocity, we implemented a Contact-assisted Invariant Extended Kalman Filter (CI-EKF) framework\cite{hartley2020contact}. This is a symmetry-aware observer design based on Lie group theory and has been shown to provide superior estimation performance compared to standard quaternion-based EKFs. The estimated velocity serves as input to the RL policy during real robot deployment.

In the training and simulation phase, to improve the transferability of the learned policy to real-world scenarios, we utilized the Humanoid Gym framework\cite{gu2024humanoid} and adopted the Isaac Gym and Mujoco platforms. The training pipeline remained consistent with the Unity-based simulation, incorporating the same feedforward control, observation space, and reward function. Domain randomization was applied throughout the training process to various physical parameters to improve the policy’s robustness to environmental variations and hardware imperfections. The specific domain randomization parameters are summarized in Table~\ref{domain_rand_params}.

\begin{figure*}
    \centering
    \includegraphics[width=0.75\textwidth]{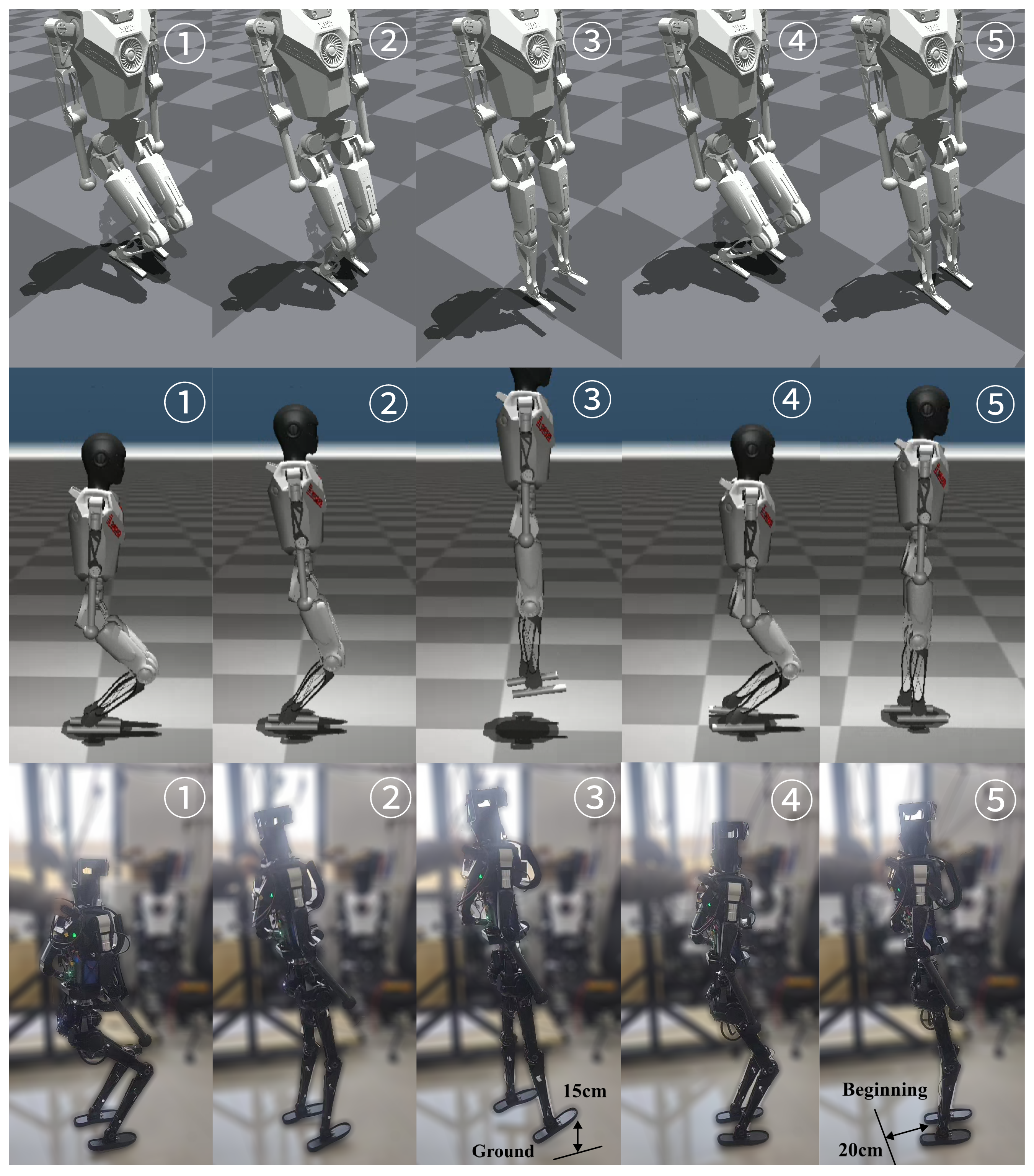}
    \caption{Performance comparison of the training policy in simulation (Isaac Gym and Mujoco) and real-world experiments, demonstrating successful transfer. The robot achieved a maximum jump height of 15\,cm, distance of 20\,cm, and peak velocity of 2\,m/s.}
    \label{simandreal}
\end{figure*}

The training process began with training the policy in Isaac Gym. We further optimized the policy by adjusting the domain randomization parameters, transferred it to MuJoCo to evaluate its performance, and then deployed it on the X02-lite robot through a sim-to-real transition. Fig. \ref{simandreal} shows the results from Isaac Gym, MuJoCo, and the real world.

In real-world tests, the robot successfully executed the jumping task, closely replicating simulation performance. During the experiments, the robot achieved a maximum jump height of 15\,cm, a single jump distance of 20\,cm, and a peak velocity of 2\,m/s.

\begin{figure*}
    \centering
    \includegraphics[width=0.5\textwidth]{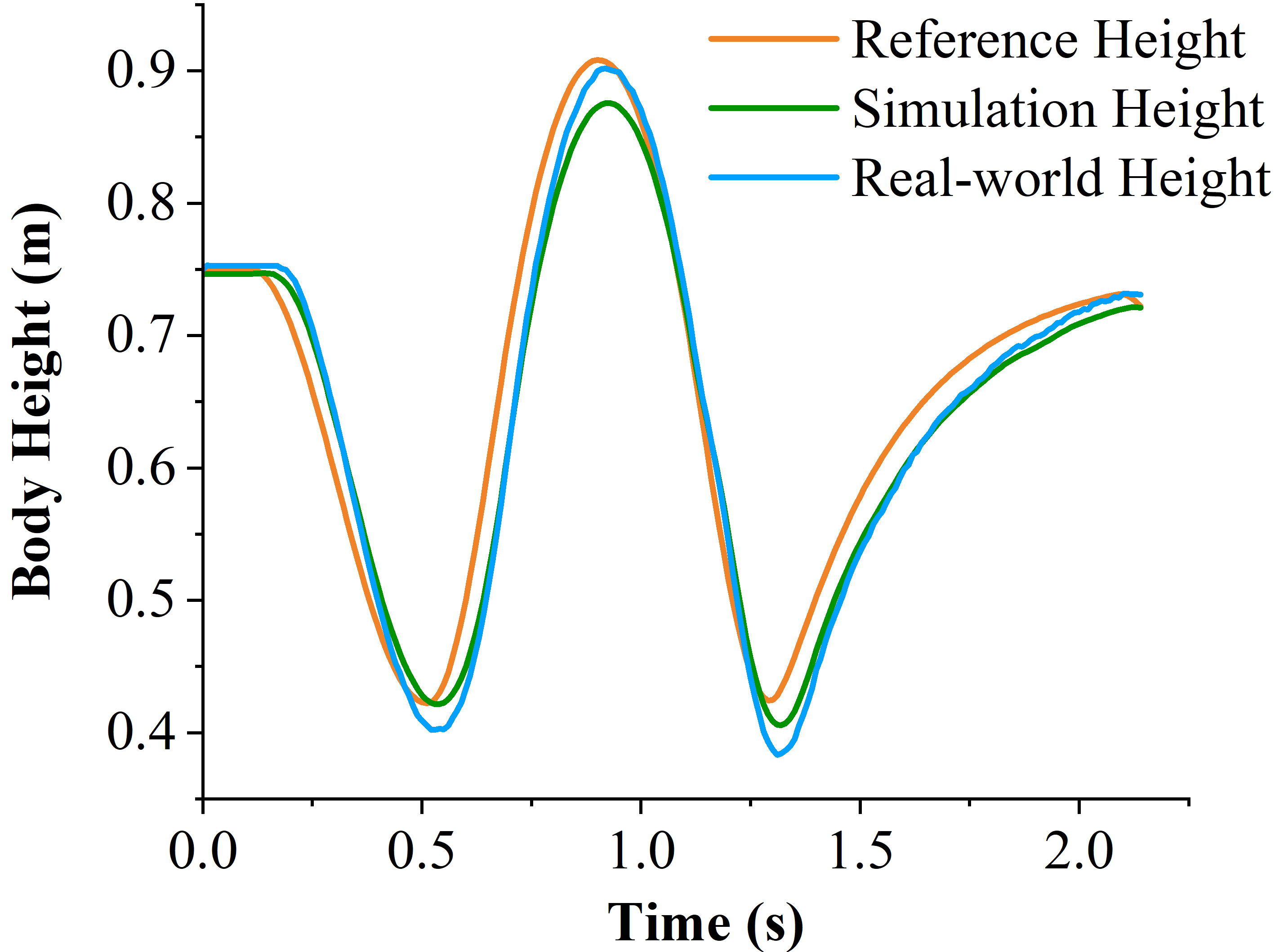} % 宽度适配跨栏，可按需调整
    \caption{Average body height comparison over one gait cycle: reference vs. simulation vs. real-world.}
    \label{x02height}
\end{figure*}

To further quantify the sim-to-real performance, we compare the robot’s average base height over a single gait cycle across three setups: reference, simulation, and real-world — as shown in Fig. \ref{x02height}. Although the overall locomotion pattern was successfully transferred, minor deviations in amplitude and phase were observed during flight and contact transitions. These discrepancies may stem from a combination of hardware limitations, sensor noise, and estimation inaccuracies — particularly during the flight phase, the linear velocity estimates provided by the CI-EKF drift, which can affect the policy’s perception of the robot’s state and lead to slight tracking errors.

\begin{table*}[htbp]
    \centering
    \small
    \caption{Average key performance metrics over one cycle under different setups: reference, simulation, and real-world.}
    \label{metrics}
    % 表格列宽适配跨栏，保持原对齐逻辑
    \begin{tabular}{@{} >{\centering\arraybackslash}p{2.5cm} >{\centering\arraybackslash}p{3cm} >{\centering\arraybackslash}p{3cm} >{\centering\arraybackslash}p{3cm} @{}}
    \toprule
    \textbf{Setup} & \textbf{Peak Velocity (m/s)} & \textbf{Jump Height (m)} & \textbf{Flight Time (s)} \\
    \midrule
    Reference  & 1.998 & 0.153 & 0.718 \\
    Simulation & 2.011 & 0.127 & 0.723 \\
    Real-world & 2.003 & 0.150 & 0.721 \\
    \bottomrule
    \end{tabular}
\end{table*}

Table~\ref{metrics} summarizes the average key performance metrics of the robot's jumps over one gait cycle under three different settings. Peak velocity exhibited nearly identical values across all setups ($\approx$ 2.0\,m/s). The simulated jump height was approximately 15\% lower than the real test (0.127\,m vs. 0.150\,m). This discrepancy may be due to limitations of the simulation environment, such as the simplified contact model. Flight time remained relatively consistent across all settings ($\approx$ 0.72\,s), indicating that the duration of the flight phase was well preserved, even in simulation.

The results suggest promising transfer performance to real-world scenarios, warranting further investigation into practical deployment.

\section{Conclusion}
This paper proposes SRL, a control framework that integrates the SLIP model with RL to optimize robot jumping performance. The SLIP model generates fundamental jumping dynamics, while  RL enhances environmental adaptability. The framework's effectiveness is validated through fixed-distance and random-distance jumping tasks on both biped and quadruped robots. Simulations and real-world experiments demonstrate centimeter-level localization accuracy (average trajectory tracking error\,$<$\,10\,cm), velocity tracking errors within ±\,3\,\% of the target values, and the quadruped's capability to ascend 0.35\,m high stairs. Due to safety and actuator constraints of the quadruped hardware platform, real-world validation on quadruped robots is not included in this work.

Current limitations include error accumulation in prolonged tasks and unstable hardware deployment. Furthermore, although SRL theoretically supports generating arbitrary 3D trajectories and then performing 3D jumping via RL, we currently limit our experiments to 2D due to hardware security concerns. Future work will focus on improving sustained stability, developing efficient sim-to-real strategies, and extending the application to 3D jumping. The proposed SRL framework offers an efficient and adaptable solution for dynamic robot control research.

\section*{Declarations}
\textbf{Funding} This work was supported by the Science and Technology Commission of Shanghai Municipality (24511103304).

\textbf{Availability of Data and Material} All relevant data are included within this paper.

\textbf{Conflict of Interest} 
The authors declare that they have no competing interests.

\textbf{Ethics Approval and Consent to Participate}  
Not applicable.

\textbf{Consent for Publication}  
Not applicable.

\printcredits

\bibliographystyle{elsarticle-num}
%\bibliographystyle{cas-model2-names}

\bibliography{cas-refs}



\end{document}